\documentclass[runningheads]{llncs}
\usepackage{graphicx}

\usepackage{hyperref}
\usepackage{amsmath}
\usepackage{amssymb}
\usepackage{booktabs}
\usepackage{float}
\usepackage{longtable}
\usepackage[ruled,vlined]{algorithm2e}
\usepackage{svg}

\SetKwFunction{FDSaturCompletion}{DSaturCompletion}
\SetKwFunction{FGISD}{GISD}
\SetKwFunction{FProjGreedyIS}{ProjGreedyIS}
\SetKwFunction{FSSLD}{SSLD}
\SetKwProg{Fn}{Function}{:}{}

\DeclareMathOperator*{\argmax}{arg\,max}
\DeclareMathOperator{\tr}{tr}
\DeclareMathOperator{\sat}{sat}
\DeclareMathOperator{\colors}{colors}

\title{One Color Preprocessing Improves DSATUR}

\author{Adam Nouira\inst{1}\orcidID{0009-0005-1741-129X} \and
Lucas Isenmann\inst{1}\orcidID{0000-0002-1460-269X}}
\authorrunning{A. Nouira and L. Isenmann}
\institute{Université de Strasbourg \\
\email{adam.nouira@etu.unistra.fr}\\
\email{lucasisenmann@unistra.fr}}

\date{}

\begin{document}
\maketitle

\section*{Abstract}

The Graph Coloring Problem (GCP) is NP-hard and DSATUR stands as one of the fastest heuristics for it despite producing colorings that typically use more colors than state-of-the-art coloring algorithms. We propose SSLD (Semidefinite Spectral Learning with DSATUR), which improves DSATUR by preprocessing a first good color class before letting DSATUR complete coloring the rest of the given graph.
We obtain this color class from a Semidefinite Programming (SDP), similar to an SDP used to compute the Lovász theta number.
To the best of our knowledge, SSLD is the first approach to improve DSATUR by preprocessing through fixed color classes.
We evaluate SSLD against DSATUR and against a naive 1-color-class preprocessing algorithm on DIMACS instances, random graphs (Erdős--Rényi, Watts-Strogatz, Barabási--Albert), Frequency Assignment and Job Shop Scheduling instances.
SSLD matches or beats DSATUR in almost every case across over 1600 benchmark instances, and out performs the naive GISD baseline, allows us to confirm the value brought by the SDP-guided choice of the first color class.
This quality comes at a runtime cost of roughly 195 times slower that DSATUR, but demonstrating that SDP-guided preprocessing of a first color class is a direction for future improvements.

\section{Introduction}

The Graph Coloring Problem (GCP) is to color the vertices of a graph using as few colors as possible such that no adjacent vertices share the same color.
The GCP can also be considered as partitioning the vertex set of the graph into a minimum number of color groups such that no vertices in each color group are adjacent.
The GCP has numerous practical applications in various domains \cite{lewis2021guide} and has been studied for a long time.
Graph coloring arises naturally in a variety of applications such as register allocation \cite{briggs1989coloring,chaitin1981register,chaitin1982register} and timetable or examination scheduling \cite{wood1969technique,berge1985graphs}.
The minimum number of colors used in proper coloring is called the chromatic number of the graph and is denoted by $\chi(G)$.
Determining the value of $\chi(G)$ is NP-hard \cite{garey2002computers} and even the $k$-Coloring problem is NP-hard for every $k \geq 3$.

Concerning exact algorithms, using dynamic programming an algorithm in $O^*(2.4423^n)$ \cite{lawler1976note} has been derived.
Using the principle of inclusion--exclusion and Yates's algorithm for the fast zeta transform, k-colorability can be decided in time $O(2^n n^{O(1)})$ \cite{bjorklund2009set} for any k.
Faster algorithms are known for 3- and 4-colorability, which can be decided in time $O(1.3289^n)$ \cite{beigel20053} and $O(1.7272^n)$ \cite{fomin2007improved} respectively.
In the case of perfect graphs, computing the chromatic number is polynomial \cite{grotschel1984polynomial}.

Nevertheless, exact algorithms are only tractable up to 300 vertices (taking 10 minutes for a random graph), that is why we need approximation algorithms or heuristics for bigger graphs.
Unfortunately, it has been shown that if certain reasonable complexity conjectures hold then $k$-Coloring is hard to approximate within $n^{1-\epsilon}$ for any $\epsilon > 0$ \cite{feige1998zero}.
Furthermore, for any constant $\gamma > 0$, there is no polynomial time algorithm that approximates the chromatic number within factor $n / 2^{(\log(n))^{3/4+\gamma}}$ where $n$ is the size of the graph assuming another reasonable complexity conjecture \cite{khot2006better}.

\subsection{Approximation algorithms}

The first approximation algorithm
shows that a version of the greedy algorithm gives an $O(n/\log n)$-approximation algorithm for $k$-coloring \cite{johnson1974worst}.

Later, \cite{karger1998approximate} introduced an approach to approximate graph coloring via semidefinite programming.
Their algorithm solves an Semidefinite Programming (SDP) relaxation linked to the Lovász theta number to obtain a vector representation of the graph, then extracts a coloring after multiple iterations of applying a random hyperplane rounding.
For a 3-colorable graph on $n$ vertices, their algorithm produces a coloring using $O(n^{0.387})$ colors, and for a $k$-colorable graph, it produces a coloring using $O(n^{1 - 3/(k+1)})$ colors.
Given that these theoretical guarantees are weaker than the $\chi$, their algorithm establishes the key connection between SDP relaxations and coloring extraction via randomized rounding that other algorithms, including the one presented in this work, build on.

\subsection{Heuristics}

Many heuristics have been proposed for graph coloring; we summarize them by family.

The earliest ones are greedy: the vertices are colored one after another, each vertex receiving a color that none of its already colored neighbours uses.
The methods of this family differ in the order in which the vertices are considered and in the way the color is chosen; DSATUR \cite{brelaz1979new} and RLF \cite{leighton1979graph} are the two best known.
They are fast, but the colorings they return use more colors than the best known ones, so they are now used mainly to produce initial solutions for other algorithms.

A second family is local search: a single coloring is modified repeatedly, by small or large steps.
Some methods keep the coloring proper at every step, others allow improper colorings and aim at reducing the number of monochromatic edges.
\cite{culberson1996exploring,joslin1999squeaky,laguna2001grasp,hertz1987using,dorne1999tabu,chams1987some,johnson1991optimization,chiarandini2002application,avanthay2003variable,morgenstern1996distributed,lewandowski1996experiments,blochliger2008graph,prestwich2002coloration,hertz2008variable}

A third family is evolutionary algorithms: a population of colorings evolves under modifications that are borrowed from local search:
\cite{fleurent1996genetic,dorne1998new,galinier1999hybrid,galinier2008adaptive,malaguti2008metaheuristic,porumbel2009diversity,lu2010memetic}.

A few methods combine several of these families \cite{sun2021solution,porumbel2010search}.
Finally, a heuristic derived from a branch-and-bound algorithm has also been proposed \cite{glover1996coloring}.

\subsection{The Lovász \texorpdfstring{$\vartheta$}{theta} Function}

Before defining the Lovász $\vartheta$ function \cite{lovasz1979shannon}, we recall some notation.
Given a graph $G$, we write $\omega(G)$ for its clique number, $\chi(G)$ for its chromatic number and $\overline{G}$ for its complement.

\paragraph{Definition.} An orthonormal representation of a graph $G=(V,E)$ with vertex set
$V = \{1, \dots, n\}$ is a set of unit vectors $(u_1, \dots, u_n)$ in $\mathbb{R}^n$ such that $u_i^T u_j = 0 \text{ if } i \neq j \text{ and } \{i,j\} \notin E$.

\paragraph{Definition (Lovász $\vartheta$ function).}
Let $G$ be a graph on $n$ vertices.
The Lovász $\vartheta$ function of $G$ is:
$$ \vartheta(G) = \min_{c, U} \max_{1 \leq i \leq n} \frac{1}{(c^T u_i)^2} $$
where the minimum is taken over all unit vectors $c$ of $\mathbb{R}^n$ and over all $U = \{u_1, \dots, u_n\}$ orthonormal representations of $G$.

The Lovász $\vartheta$ function is upper bounded by the chromatic number of the complement graph:

\paragraph{Theorem \cite{lovasz1979shannon}.} $\omega(G) \leq \vartheta(\overline{G}) \leq \chi(G)$.

\subsection{Semidefinite Programming Formulation}

We write $S^n$ for the space of real symmetric $n \times n$ matrices,
$(A, B) = \operatorname{tr}(AB)$ for the trace inner product on $S^n$, and
$A \succeq 0$ to say that $A \in S^n$ is positive semidefinite.

A semidefinite program consists in optimizing a linear function over the
intersection of the cone of positive semidefinite matrices with an affine
subspace.
It contains linear programming as the particular case where all the
matrices involved are diagonal.
SDP can be solved in polynomial time to arbitrary precision \cite{grotschel1981ellipsoid}.

The Lovász $\vartheta$ function $\vartheta(G)$ can be computed thanks to the following SDP formulation, which we call the L-SDP of $G$:

\begin{align*}
\text{maximize } & X_{n+2,n+2} \\
\text{where } & X \succeq 0 \text{ and } X \in S^{n+2}(\mathbb{R}) \\
\text{subject to } & X_{i,i} = 1, \quad \forall\, 1 \leq i \leq n+1 \\
& X_{i,j} = 0 \text{ if } \{i,j\} \notin E \\
& X_{n+1,j} \geq X_{n+2,n+2}, \quad \forall\, 1 \leq j \leq n \\
& X_{n+2,j} = 0, \quad \forall\, 1 \leq j \leq n+1.
\end{align*}

According to \cite{lovasz1979shannon}, by calling $t$ the optimal value $X_{n+2,n+2}$, then we have $\vartheta(G) = 1/\sqrt{t}$.
Thus $\vartheta(G)$ can be computed in polynomial time.
Note that the orthonormal representation corresponding to this optimal value can be derived from the Gram representation of $X^*$.

\subsection{Our Contribution}

We focus on the greedy algorithm DSATUR which is one of the fastest heuristics.
The goal is to improve the number of colors returned by this algorithm by starting with a good independent set (a color class) and complementing the color class with the DSATUR algorithm.
Our SSLD algorithm will try different independent sets and return the best result.

To find appropriate independent sets, our algorithm will make use of a Semidefinite Programming (SDP) relaxation with a spectral decomposition and randomized rounding procedure.
We propose a new SDP similar to the one used in the Lovász Theta function which is a parameter lower bound on the chromatic number.

We conduct an experimental assessment on benchmarks from the DIMACS competition, randomly generated graphs, frequency assignment problems, and job shop scheduling instances.
We show that SSLD consistently uses fewer colors than DSATUR on these instances.

To the best of our knowledge, our algorithm is the first which tries to improve the DSATUR by preprocessing by fixing some color classes.

\section{DSATUR preprocessing}\label{sec:dsatur_preprocessing}

Given any algorithm $\mathcal{A}$ that returns an independent set $I$ of $G$, the preprocessing strategy assigns color $0$ to all vertices in $I$ and completes the coloring of the remaining vertices with DSATUR (Algorithm~\ref{alg-dsatur-completion}).
The number of colors used in the final coloring depends on the size of $I$ and on which vertices it contains because a poorly chosen independent set, even a large one, can leave a harder subgraph for DSATUR and result in more colors than plain DSATUR.

\begin{algorithm}[h]
\caption{DSATUR Completion}\label{alg-dsatur-completion}
\Fn{\FDSaturCompletion{$G(V,E)$, $c : V' \to \mathbb{N}$ (partial coloring)}}{
  $U \leftarrow V \setminus V'$\;
  \While{$U \neq \emptyset$}{
    $v \leftarrow \displaystyle\argmax_{u \in U} (|\sat(u)|, \deg(u))$\;
    $c(v) \leftarrow \min \{ k \in \mathbb{N} : k \notin c(N(v)) \}$\;
    $U \leftarrow U \setminus \{v\}$\;
  }
  \KwRet{$c$}
}
\end{algorithm}

\section{Naive DSATUR preprocessing}\label{sec:gisd}

We initially tried the most naive possible variant of our idea: preprocess DSATUR by fixing one color class taken as a maximal independent set built with a greedy algorithm, and completing the rest with DSATUR as before.
This algorithm is described in Algorithm~\ref{algo:gisd} and we call it Greedy Independent Set with DSATUR (GISD).

Since it requires no SDP solve, its overhead over plain DSATUR is negligible, as reflected by its near-zero $t(s)$ column throughout (Section~\ref{sec:exp}).

\begin{algorithm}[h]
\caption{GISwDSATUR}\label{algo:gisd}
\Fn{\FGISD{$G(V,E)$}}{
  $I \leftarrow \emptyset$\;
  \For{$v \in V$}{
    \If{$v$ has no neighbor in $I$}{
      $I \leftarrow I \cup \{v\}$\;
    }
  }
  $c \leftarrow \{v : 0 \mid v \in I\}$\;
  $c \leftarrow$ \FDSaturCompletion{$G, c$}\;
  \KwRet{$|\colors(c)|$}
}
\end{algorithm}

\section{SSLD Description}\label{sec:ssld}

\subsection{The D-SDP}\label{sec:dsdp}

We define $J \in \mathbb{R}^{n \times n}$ as the all-ones matrix. The SDP solved by SSLD is:

\begin{align*}
\operatorname*{maximize}_{X} \quad & \tr(XJ) \\
\text{subject to} \quad & X \succeq 0 \\
& X_{ii} = 1 \quad \forall i \in V \\
& X_{ij} = 0 \quad \forall ij \in E
\end{align*}
This problem computes an orthonormal representation of $\overline{G}$ and is related to the SDP used by Lovász to compute $\vartheta(G)$, but differs in its objective and carries no proven theoretical guarantees.

Orthogonal representations will be used as follows to detect color classes.
If we consider a unit vector called, a centroid, and if we compute the scalar products of the vertices with this centroid, then all the vertices which have a high scalar products are not adjacent (because adjacent vertices are orthogonal and thus cannot be in a small cap).
Therefore we will try to find good centroids for detecting color classes.
This algorithm is described in Algorithm~\ref{algo:pgis}.

Algorithm~\ref{algo:pgis} generalizes GISD (Algorithm~\ref{algo:gisd}): GISD is the special case obtained by replacing the spectral score $p$ with the identity ordering on $V$. As we previously said, since it requires no SDP solve, its overhead over plain DSATUR is negligible. However, as we show in Section~\ref{sec:exp}, the naive preprocessing is inconsistent: it sometimes matches or slightly improves on DSATUR, but on several instances (e.g. \texttt{DSJC125.9}, \texttt{queen11\_11}) it performs strictly worse, since an arbitrarily-ordered independent set is not guaranteed to be a good first color class. SSLD addresses this by replacing the identity ordering with one derived from the SDP relaxation.

\begin{algorithm}[h]
\caption{Projection-Greedy Independent Set}\label{algo:pgis}
\Fn{\FProjGreedyIS{$G(V,E)$, $S \subseteq V$, $p \in \mathbb{R}^n$}}{
  $I \leftarrow \emptyset$\;
  \For{$v \in S$ sorted by decreasing $p_v$}{
    \If{$v$ has no neighbor in $I$}{
      $I \leftarrow I \cup \{v\}$\;
    }
  }
  \KwRet{$I$}
}
\end{algorithm}

The Lovász SDP is also finding an orthogonal representation of $G$ but it looks for one maximizing the minimum of the scalar products with a unit vector (called the handle).
Thus this optimal orthogonal representation keeps the vertices in the smallest cap.
Thus the color classes centroids will get more vertices because the vertices are in a small cap.
And thus there may not be many color classes.

The objective of the D-SDP is
$$ \tr(XJ) = \sum_{i \in V} X_{ii} + \sum_{ij \in E} (X_{ij} + X_{ji}) + \sum_{i<j,\, ij \notin E} 2 X_{ij} = n + 2 \sum_{i<j,\, ij \notin E} \langle v_i, v_j \rangle $$
where the $v_i$ is the vector representing the vertex $i$.
So the D-SDP directly rewards aligning vertices that could form a common independent set.
Therefore the D-SDP may reward big independent sets.

Nevertheless, the Lovász SDP is slower to compute and gives no better results according to Table~\ref{table:versus-lovasz}.

\subsection{Algorithm Description}\label{sec:algo}

SSLD solves the D-SDP to obtain a matrix $X^*$, whose eigendecomposition $X^* = Q \Lambda Q^\top$ results in a spectral embedding where each vertex $i$ is represented as a vector $V_i \in \mathbb{R}^n$ on the unit sphere.
At each of $T$ iterations, SSLD draws a random direction $r$ according to $\mathcal{N}(0, I)$ and projects all vertex embeddings onto it, producing scalar scores $p = Vr$.
A greedy independent set is then extracted from the full vertex set by processing vertices in decreasing order of $p_v$.
This independent set is assigned color 0, and the remaining uncolored subgraph is completed with the DSATUR heuristic.
The coloring achieving the fewest colors across all $T$ attempts is returned.

\begin{algorithm}[h]
\caption{Semidefinite Spectral Learning with DSATUR}
\Fn{\FSSLD{a graph $G$}}{
  $X^* \leftarrow \mathrm{SDP}(G)$\;
  $(\Lambda, Q) \leftarrow$ eigendecomposition of $X^*$\;
  $V \leftarrow Q \cdot \mathrm{diag}(\sqrt{\max(\Lambda, 0)})$\;
  $k^* \leftarrow +\infty$\;
  \For{$i = 1$ \KwTo $T$}{
    $r \leftarrow \mathcal{N}(0, I)$\;
    $p \leftarrow Vr$\;
    $I \leftarrow$ \FProjGreedyIS{$G, V, p$}\;
    $c \leftarrow$ \FDSaturCompletion{$G, \{v : 0 \mid v \in I\}$}\;
    \If{$|\colors(c)| < k^*$}{
      $k^* \leftarrow |\colors(c)|$\;
    }
  }
  \KwRet{$k^*$}
}
\end{algorithm}

\subsubsection{Parameter Selection}\label{sec:param}

\begin{figure}[h]
\centering
\includegraphics[width=0.8\textwidth]{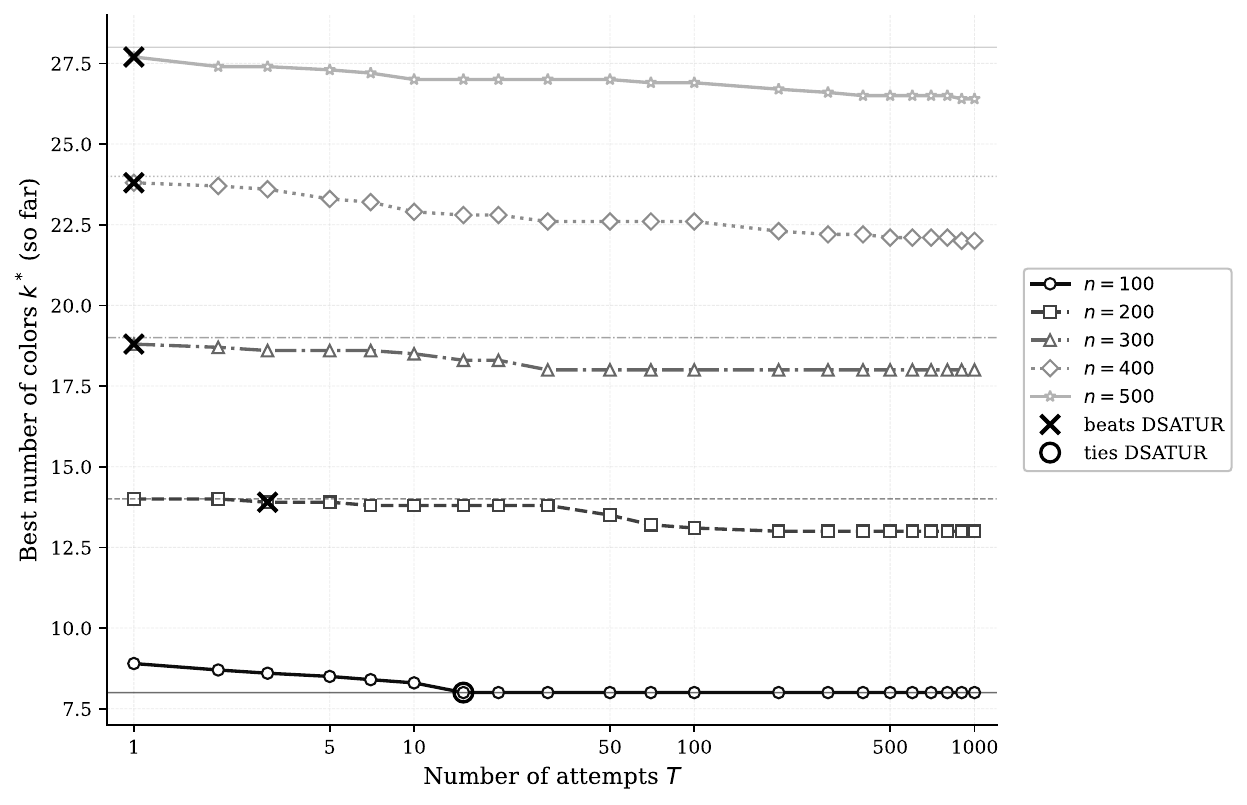}
\caption{SSLD convergence vs. number of attempts.}
\label{ssldT}
\end{figure}

The number of attempts $T$ controls the trade-off between solution quality and runtime. To quantify this trade-off, we generate Erdős--Rényi graphs $G(n, 0.2)$ for $n \in \{100, 200, 300, 400, 500\}$ and, for each graph, run SSLD over 10 independent random seeds, recording the best chromatic number found so far at each attempt count up to $T = 1000$. Figure~\ref{ssldT} reports the mean of these best-so-far values as a function of $T$, together with the chromatic number obtained by plain DSATUR (without SDP preprocessing) as a horizontal reference for each $n$. An ``x'' marks the first attempt at which SSLD strictly improves on the DSATUR baseline, and a ``o'' marks an instance where SSLD ties but does not surpass it. Across all tested sizes, the bulk of the improvement is obtained within the first few dozen attempts, after which the curves plateau, suggesting that $T$ can be fixed to a modest value without a significant loss in solution quality.

We fix $T = 500$ as the default number of attempts, a choice supported across both the sparse regime ($p = 0.2$, $n \leq 500$, Figure~\ref{ssldT}) and denser instances ($p \in \{0.2, 0.4, 0.6\}$, $n \in \{100, \dots, 500\}$): in every configuration tested, the mean best (so far) chromatic number at $T = 500$ differs from that at $T = 1000$ by at most $0.2$. Therefore, doubling the attempt budget beyond $500$ would yield no meaningful gain in coloring quality regardless of density, since each attempt adds a fixed cost (a random projection, greedy independent set extraction, and DSATUR completion) beyond the single SDP solve per graph, $T = 500$ captures nearly all of the achievable improvement over DSATUR.

Note that we can also run SSLD without a fixed T. It would be a time-budget variant of the algorithm that would run until a given time limit is reached, making it more flexible.

\section{Experimentation}\label{sec:exp}

\subsection{Benchmark Instances}\label{sec:benchmarks}

We evaluate SSLD on different categories of instances.
The DIMACS challenge \cite{johnson1996cliques} provides the standard hard instances used to compare graph coloring algorithms in the literature.
We complement these with three families of synthetic random graphs: Erdős--Rényi \cite{erdos1959random}, Watts-Strogatz \cite{watts1998collective}, and Barabási--Albert \cite{barabasi1999emergence}.
We further evaluate on real-world instances derived from two applications: frequency assignment problems and job shop scheduling.
Finally, we evaluate on the adversarial family of graphs of Spinrad and Vijayan \cite{spinrad1985worst}, for which DSATUR provably requires $n$ colors on a 3-colorable graph, providing a worst-case stress test for any DSATUR-based method.

\subsubsection{Hard-to-Color Instances for DSATUR}

We usually evaluate the performance of greedy coloring heuristics through their worst-case asymptotic guarantees.
Spinrad and Vijayan \cite{spinrad1985worst} demonstrated that the DSATUR algorithm admits an adversarial family $G_n$ of 3-colorable graphs on $3n - 4$ vertices for which DSATUR may use $n$ colors.

For any $n \in \mathbb{N}_{\geq 3}$, $G_n = (V_n, E_n)$ where $V_n = W_n \cup W'_n \cup W''_n$ and $E_n = E^1_n \cup E^2_n \cup E^3_n$,
with
$W_n = \{v_1, \dots, v_{n-2}\}$,
$W'_n = \{v'_1, \dots, v'_{n-1}\}$,
$W''_n = \{v''_2, \dots, v''_n\}$,
and edge sets
$E^1_n = \{(v'_i, v''_{i+1}) : 0 < i < n\}$, $E^2_n = \{(v_i, v'_j) : i \neq j\}$, $E^3_n = \{(v_i, v''_j) : i < j\}$.

\begin{figure}[h]
\centering
\includegraphics[width=0.9\textwidth]{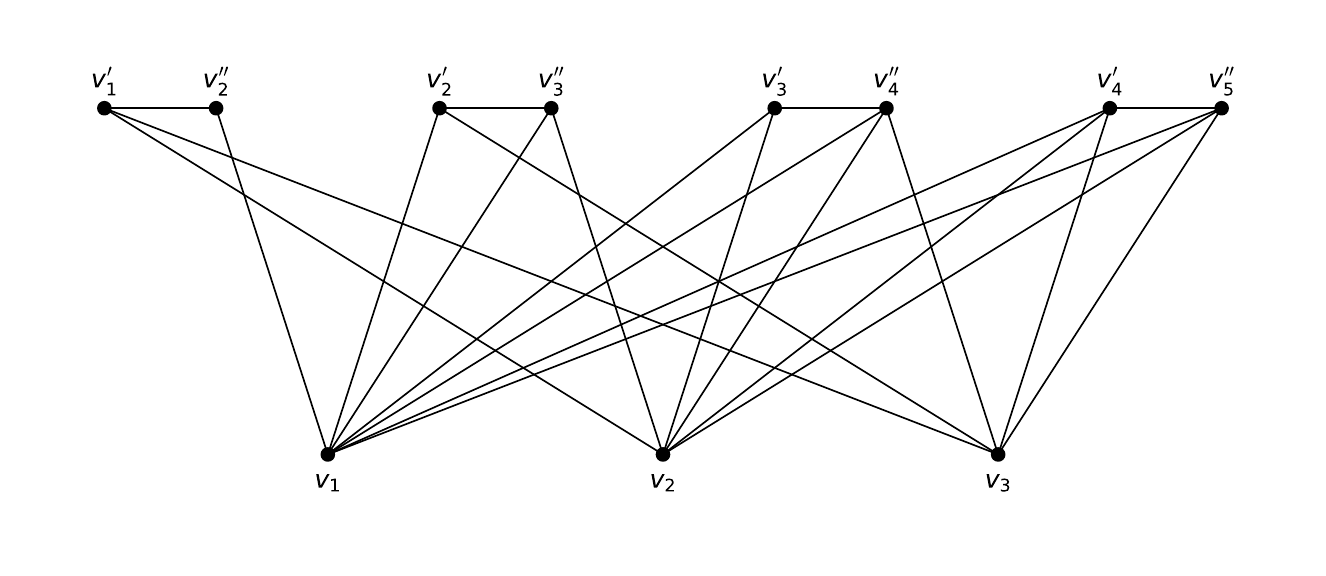}
\caption{The graph $G_5$ for which DSATUR use $5$ colors despite $\chi(G_5) = 3$.}
\label{SVGraph}
\end{figure}

Since there are no edges between vertices within the same partition, $G_n$ is 3-colorable, with $W_n$, $W'_n$, and $W''_n$ forming the three independent color
classes (Figure~\ref{SVGraph}).
Spinrad and Vijayan show that the adversarial coloring sequence $s_n$ causes DSATUR to assign a distinct color to each of $v_i$, $v'_i$, and $v''_i$, resulting in $n$ colors on a graph with chromatic number 3.
This family thus provides a worst-case benchmark against which the improvement brought by SSLD can be measured.

\subsubsection{DIMACS Instances}

The DIMACS implementation challenge \cite{johnson1996cliques} established a standard set of benchmark instances for the graph coloring problem, now widely used to evaluate and compare coloring algorithms in the literature.
The benchmark contains instances from several structured families, each with distinct combinatorial properties that stress different aspects of coloring algorithms.

The \texttt{DSJC} family consists of random graphs $G(n, p)$ generated with the DSJC generator, parameterized by density $p \in \{0.1, 0.5, 0.9\}$ and size $n \in \{125, 250, 500, 1000\}$.
The \texttt{flat} family contains equitably $k$-colorable random graphs, designed so that the chromatic number is known exactly.
The \texttt{queen} family derives from the $n \times n$ queens problem on a chessboard where vertices represent squares and edges connect squares that share a row, column, or diagonal.
The \texttt{myciel} family consists of Mycielski graphs $M_k$, which are triangle-free graphs with arbitrarily large chromatic number, constructed recursively so that $\chi(M_k) = k$ while $\omega(M_k) = 2$.
The \texttt{miles} family encodes geometric distance graphs on US cities.
Finally, the register allocation family provides conflict graphs derived from compiler register allocation problems.

We evaluate over these families covering sparse, dense, structured, and random instances, following a subset of the selection in \cite{malaguti2008metaheuristic}.

\subsubsection{Scheduling Instances}

The job shop scheduling problem consists of assigning a set of jobs to machines over a set of time slots subject to precedence and resource constraints.
A classical reduction to graph coloring \cite{berge1985graphs} assigns a vertex to each job and introduces an edge between two jobs that cannot be executed simultaneously due to shared resource requirements.
A proper $k$-coloring of the resulting conflict graph yields a feasible schedule using $k$ time slots, and the chromatic number $\chi(G)$ equals the minimum number of time slots required. We evaluate on instances from the COLOR benchmark \cite{johnson1996cliques}, which provide conflict graphs derived from real industrial job shop scheduling problems.

\subsubsection{Frequency Assignment Problems (FAP)}

The frequency assignment problem (FAP) arises in the planning of radio communication networks, where a set of transmitters must be assigned operating frequencies such that co-channel interference between geographically proximate transmitters is avoided.

Formally, given a set of transmitters $T$ and an interference relation $I \subseteq \binom{T}{2}$ consisting of the unordered pairs of transmitters that mutually interfere, the problem requires finding an assignment $f : T \to \mathbb{N}$ such that $f(t_i) \neq f(t_j)$ for all $\{t_i, t_j\} \in I$, while minimizing $|f(T)|$. This is equivalent to the chromatic number of the interference graph $G = (T, I)$.

We use the CELAR and GRAPH instance families from the Radio Link Frequency Assignment Problem benchmark \cite{dorne1999tabu}, derived from French military communication network planning data and widely adopted as standard benchmarks in the graph coloring literature \cite{malaguti2008metaheuristic,sun2021solution}.

\subsubsection{Random Graphs}

For random graphs, we have chosen these three most common models:
\begin{itemize}
\item Erdős--Rényi model \cite{erdos1959random} (also known as $G(n,p)$): given parameters $n \in \mathbb{N}$ and $p \in [0,1]$, creates a graph with $n$ vertices such that each edge appears independently with probability $p$.
\item Watts-Strogatz model \cite{watts1998collective}: given parameters $n \in \mathbb{N}$, $k \in \mathbb{N}$ and $\beta \in [0,1]$, this algorithm creates a graph with $n$ vertices by rewiring a ring lattice, producing graphs whose structural properties resemble those found in real-world networks.
\item Barabási--Albert model \cite{barabasi1999emergence}: given parameters $n \in \mathbb{N}$ and $m \in \mathbb{N}$, this algorithm grows a graph to $n$ vertices by adding one vertex at a time, each new vertex connecting to $m$ existing vertices chosen with probability proportional to their current degree, a preferential attachment mechanism where already well-connected vertices tend to attract even more connections.
\end{itemize}

\subsection{Results}\label{sec:results}

We use the implementation of DSATUR from networkx Python package.
Experiments are conducted under the following settings:

\begin{center}
\begin{tabular}{ll}
\toprule
\textbf{OS} & Windows 11 Pro 22H2 (build 22621.382) \\
\textbf{CPU} & Intel Core i7-14700K @ 5.60 GHz \\
\textbf{RAM} & 16 GB DDR5 6000 MHz CL40 \\
\textbf{Language} & Python 3.13.5 \\
\textbf{Libraries} & NumPy, SciPy, NetworkX, CVXPY \\
\textbf{SDP solver} & SCS (via CVXPY, tolerance $\epsilon = 10^{-6}$) \\
\bottomrule
\end{tabular}
\end{center}

Each SSLD result is reported over 5 independent runs with different random seeds. The hit column indicates how many of those 5 runs achieved the reported number of colors k, giving a sense of the algorithm.

\subsubsection{Lovász vs. D-SDP: preliminary comparison}

\begin{table}[H]
\centering
\footnotesize
\begin{tabular}{lrrrrrrrr}
\toprule
& & & \multicolumn{3}{c}{\textbf{SSLD [D-SDP]}} & \multicolumn{3}{c}{\textbf{SSLD [L-SDP]}} \\
\cmidrule(lr){4-6} \cmidrule(lr){7-9}
\textbf{Instance} & $|V|$ & $|E|$ & $k$ & hit & $t(s)$ & $k$ & hit & $t(s)$ \\
\midrule
\multicolumn{9}{l}{\textit{DSJC}} \\
DSJC125.1 & 125 & 736 & \textbf{6} & 5/5 & 1.576 & \textbf{6} & 5/5 & 3.751 \\
DSJC125.5 & 125 & 3891 & \textbf{19} & 1/5 & 1.728 & \textbf{19} & 1/5 & 7.588 \\
DSJC125.9 & 125 & 6961 & \textbf{48} & 4/5 & 2.503 & \textbf{48} & 5/5 & 4.976 \\
DSJC250.1 & 250 & 3218 & \textbf{9} & 5/5 & 7.619 & \textbf{9} & 4/5 & 21.1 \\
DSJC250.5 & 250 & 15668 & \textbf{34} & 1/5 & 7.453 & 35 & 5/5 & 38.3 \\
\midrule
\multicolumn{9}{l}{\textit{R}} \\
R125.1 & 125 & 209 & \textbf{5} & 5/5 & 3.672 & \textbf{5} & 5/5 & 3.482 \\
R125.1c & 125 & 7501 & \textbf{46} & 5/5 & 11.0 & \textbf{46} & 5/5 & 2.746 \\
R125.5 & 125 & 3838 & \textbf{37} & 5/5 & 14.3 & \textbf{37} & 5/5 & 126.8 \\
R250.1 & 250 & 867 & \textbf{8} & 5/5 & 38.6 & \textbf{8} & 5/5 & 104.0 \\
\midrule
\multicolumn{9}{l}{\textit{flat}} \\
flat300\_20\_0 & 300 & 21375 & \textbf{39} & 5/5 & 11.0 & \textbf{39} & 5/5 & 101.7 \\
flat300\_26\_0 & 300 & 21633 & \textbf{39} & 1/5 & 11.3 & \textbf{39} & 1/5 & 60.5 \\
flat300\_28\_0 & 300 & 21695 & \textbf{39} & 4/5 & 11.0 & \textbf{39} & 5/5 & 69.5 \\
\bottomrule
\end{tabular}
\caption{}
\label{table:versus-lovasz}
\end{table}

\subsubsection{Hard-to-Color Instances benchmark}

\begin{table}[H]
\centering
\footnotesize
\begin{tabular}{lrrrrrrrr}
\toprule
& & & \multicolumn{2}{c}{\textbf{DSATUR}} & \multicolumn{2}{c}{\textbf{GISD}} & \multicolumn{2}{c}{\textbf{SSLD}} \\
\cmidrule(lr){4-5} \cmidrule(lr){6-7} \cmidrule(lr){8-9}
\textbf{Instance} & $|V|$ & $\chi$ & $k$ & $t(s)$ & $k$ & $t(s)$ & $k$ & $t(s)$ \\
& & & & & & & (hit) & \\
\midrule
$G_5$ & 31 & 3 & 5 & 0.000 & \textbf{3} & 0.000 & \textbf{3} (5/5) & 0.077 \\
$G_{10}$ & 66 & 3 & 10 & 0.002 & \textbf{3} & 0.000 & \textbf{3} (5/5) & 0.339 \\
$G_{20}$ & 136 & 3 & 20 & 0.011 & \textbf{3} & 0.001 & \textbf{3} (5/5) & 2.173 \\
$G_{50}$ & 346 & 3 & 50 & 0.126 & \textbf{3} & 0.001 & \textbf{3} (5/5) & 20.5 \\
$G_{100}$ & 696 & 3 & 100 & 0.952 & \textbf{3} & 0.005 & \textbf{3} (5/5) & 102.9 \\
\bottomrule
\end{tabular}
\caption{}
\end{table}

\subsubsection{DIMACS benchmark}

{\footnotesize
\begin{longtable}{lrrrrrrrr}
\caption{DIMACS benchmark results.}\label{table:dimacs-benchmark} \\
\toprule
& & & \multicolumn{2}{c}{\textbf{DSATUR}} & \multicolumn{2}{c}{\textbf{GISD}} & \multicolumn{2}{c}{\textbf{SSLD}} \\
\cmidrule(lr){4-5} \cmidrule(lr){6-7} \cmidrule(lr){8-9}
\textbf{Instance} & $|V|$ & $|E|$ & $k$ & $t(s)$ & $k$ & $t(s)$ & $k$ (hit) & $t(s)$ \\
\midrule
\endfirsthead
\multicolumn{9}{l}{\small\textit{(continued from previous page)}} \\
\toprule
& & & \multicolumn{2}{c}{\textbf{DSATUR}} & \multicolumn{2}{c}{\textbf{GISD}} & \multicolumn{2}{c}{\textbf{SSLD}} \\
\cmidrule(lr){4-5} \cmidrule(lr){6-7} \cmidrule(lr){8-9}
\textbf{Instance} & $|V|$ & $|E|$ & $k$ & $t(s)$ & $k$ & $t(s)$ & $k$ (hit) & $t(s)$ \\
\midrule
\endhead
\midrule
\multicolumn{9}{r}{\small\textit{continued on next page}} \\
\endfoot
\bottomrule
\endlastfoot
\multicolumn{9}{l}{\textit{DSJC}} \\
DSJC125.1 & 125 & 736 & \textbf{6} & 0.007 & \textbf{6} & 0.001 & \textbf{6} (5/5) & 1.576 \\
DSJC125.5 & 125 & 3891 & 22 & 0.016 & 22 & 0.002 & \textbf{19} (1/5) & 1.728 \\
DSJC125.9 & 125 & 6961 & 51 & 0.025 & 53 & 0.002 & \textbf{48} (4/5) & 2.503 \\
DSJC250.1 & 250 & 3218 & 10 & 0.036 & 10 & 0.003 & \textbf{9} (5/5) & 7.619 \\
DSJC250.5 & 250 & 15668 & 37 & 0.104 & 37 & 0.007 & \textbf{34} (1/5) & 7.453 \\
DSJC250.9 & 250 & 27897 & 92 & 0.165 & 89 & 0.008 & \textbf{85} (3/5) & 12.4 \\
DSJC500.1 & 500 & 12458 & 16 & 0.232 & 16 & 0.014 & \textbf{15} (5/5) & 30.9 \\
DSJC500.5 & 500 & 62624 & 65 & 0.816 & 65 & 0.026 & \textbf{62} (2/5) & 37.1 \\
DSJC500.9 & 500 & 112437 & 170 & 1.384 & 165 & 0.031 & \textbf{159} (1/5) & 80.1 \\
DSJC1000.1 & 1000 & 49629 & 27 & 1.709 & 26 & 0.056 & \textbf{25} (5/5) & 131.0 \\
DSJC1000.5 & 1000 & 249826 & 115 & 7.328 & 113 & 0.098 & \textbf{112} (5/5) & 128.1 \\
DSJC1000.9 & 1000 & 449449 & 299 & 12.574 & 305 & 0.131 & \textbf{292} (2/5) & 258.3 \\
\midrule
\multicolumn{9}{l}{\textit{DSJR}} \\
DSJR500.1 & 500 & 3555 & \textbf{13} & 0.108 & 14 & 0.010 & T/O (--) & 3600 \\
DSJR500.1c & 500 & 121275 & \textbf{90} & 1.558 & \textbf{90} & 0.032 & T/O (--) & 3600 \\
\midrule
\multicolumn{9}{l}{\textit{R}} \\
R125.1 & 125 & 209 & \textbf{5} & 0.005 & \textbf{5} & 0.000 & \textbf{5} (5/5) & 3.672 \\
R125.1c & 125 & 7501 & \textbf{46} & 0.025 & \textbf{46} & 0.002 & \textbf{46} (5/5) & 11.0 \\
R125.5 & 125 & 3838 & 38 & 0.017 & 38 & 0.002 & \textbf{37} (5/5) & 14.3 \\
R250.1 & 250 & 867 & \textbf{8} & 0.022 & \textbf{8} & 0.002 & \textbf{8} (5/5) & 38.6 \\
R250.1c & 250 & 30227 & 65 & 0.187 & 65 & 0.008 & \textbf{64} (5/5) & 210.3 \\
R250.5 & 250 & 14849 & 68 & 0.109 & 68 & 0.006 & \textbf{66} (2/5) & 290.9 \\
\midrule
\multicolumn{9}{l}{\textit{Rg}} \\
R50\_1g & 50 & 108 & 4 & 0.001 & 4 & 0.000 & \textbf{3} (5/5) & 0.166 \\
R50\_5g & 50 & 612 & 11 & 0.002 & 11 & 0.000 & \textbf{10} (5/5) & 0.217 \\
R50\_9g & 50 & 1092 & 22 & 0.002 & 22 & 0.000 & \textbf{21} (5/5) & 0.439 \\
R75\_1g & 70 & 251 & 5 & 0.002 & 5 & 0.000 & \textbf{4} (5/5) & 0.328 \\
R75\_5g & 75 & 1407 & 15 & 0.004 & 16 & 0.001 & \textbf{14} (5/5) & 0.457 \\
R75\_9g & 75 & 2513 & 36 & 0.006 & 35 & 0.001 & \textbf{34} (5/5) & 0.816 \\
R100\_1g & 100 & 509 & 6 & 0.004 & 6 & 0.000 & \textbf{5} (5/5) & 0.809 \\
R100\_5g & 100 & 2456 & 18 & 0.008 & \textbf{17} & 0.001 & \textbf{17} (5/5) & 0.934 \\
R100\_9g & 100 & 4438 & 41 & 0.012 & 41 & 0.001 & \textbf{38} (5/5) & 1.339 \\
\midrule
\multicolumn{9}{l}{\textit{flat}} \\
flat300\_20\_0 & 300 & 21375 & 42 & 0.178 & 40 & 0.008 & \textbf{39} (5/5) & 11.0 \\
flat300\_26\_0 & 300 & 21633 & 41 & 0.168 & 43 & 0.008 & \textbf{39} (1/5) & 11.3 \\
flat300\_28\_0 & 300 & 21695 & 42 & 0.167 & 42 & 0.008 & \textbf{39} (4/5) & 11.0 \\
flat1000\_50\_0 & 1000 & 245000 & 114 & 6.870 & 115 & 0.097 & \textbf{111} (4/5) & 119.9 \\
flat1000\_60\_0 & 1000 & 245830 & 114 & 6.919 & 114 & 0.100 & \textbf{110} (1/5) & 118.7 \\
flat1000\_76\_0 & 1000 & 246708 & 115 & 7.209 & 115 & 0.100 & \textbf{111} (1/5) & 128.0 \\
\midrule
\multicolumn{9}{l}{\textit{queen}} \\
queen5\_5 & 25 & 160 & \textbf{5} & 0.000 & \textbf{5} & 0.000 & \textbf{5} (5/5) & 0.086 \\
queen6\_6 & 36 & 290 & 9 & 0.001 & 9 & 0.000 & \textbf{8} (5/5) & 0.220 \\
queen7\_7 & 49 & 476 & 11 & 0.001 & 10 & 0.000 & \textbf{9} (5/5) & 0.587 \\
queen8\_8 & 64 & 728 & 12 & 0.002 & 11 & 0.000 & \textbf{10} (5/5) & 0.541 \\
queen8\_12 & 96 & 1368 & 14 & 0.006 & 15 & 0.001 & \textbf{12} (1/5) & 0.857 \\
queen9\_9 & 81 & 1056 & 13 & 0.004 & 14 & 0.001 & \textbf{11} (5/5) & 0.987 \\
queen10\_10 & 100 & 1470 & 14 & 0.006 & 13 & 0.001 & \textbf{12} (1/5) & 1.394 \\
queen11\_11 & 121 & 1980 & 15 & 0.009 & 17 & 0.001 & \textbf{14} (5/5) & 2.491 \\
queen12\_12 & 144 & 2596 & 16 & 0.014 & 16 & 0.002 & \textbf{15} (5/5) & 3.718 \\
queen13\_13 & 169 & 3328 & 17 & 0.031 & 18 & 0.002 & \textbf{16} (3/5) & 4.712 \\
queen14\_14 & 196 & 4186 & 19 & 0.030 & 20 & 0.003 & \textbf{17} (2/5) & 7.667 \\
queen15\_15 & 225 & 5180 & 21 & 0.040 & 21 & 0.003 & \textbf{19} (5/5) & 11.0 \\
queen16\_16 & 256 & 6320 & 23 & 0.056 & 21 & 0.004 & \textbf{20} (5/5) & 14.4 \\
\midrule
\multicolumn{9}{l}{\textit{myciel}} \\
myciel3 & 11 & 20 & \textbf{4} & 0.000 & \textbf{4} & 0.000 & \textbf{4} (5/5) & 0.024 \\
myciel4 & 23 & 71 & \textbf{5} & 0.000 & \textbf{5} & 0.000 & \textbf{5} (5/5) & 0.076 \\
myciel5 & 47 & 236 & \textbf{6} & 0.001 & \textbf{6} & 0.000 & \textbf{6} (5/5) & 0.403 \\
myciel6 & 95 & 755 & \textbf{7} & 0.005 & \textbf{7} & 0.000 & \textbf{7} (5/5) & 63.6 \\
myciel7 & 191 & 2360 & \textbf{8} & 0.024 & \textbf{8} & 0.001 & \textbf{8} (5/5) & 1468.9 \\
\midrule
\multicolumn{9}{l}{\textit{miles}} \\
miles250 & 128 & 387 & \textbf{8} & 0.006 & \textbf{8} & 0.001 & \textbf{8} (5/5) & 23.1 \\
miles500 & 128 & 1170 & \textbf{20} & 0.009 & \textbf{20} & 0.001 & \textbf{20} (5/5) & 8.289 \\
miles750 & 128 & 2113 & \textbf{31} & 0.012 & \textbf{31} & 0.001 & \textbf{31} (5/5) & 185.6 \\
\midrule
\multicolumn{9}{l}{\textit{mug}} \\
mug88\_1 & 88 & 146 & \textbf{4} & 0.002 & \textbf{4} & 0.000 & \textbf{4} (5/5) & 17.2 \\
mug88\_25 & 88 & 146 & \textbf{4} & 0.002 & \textbf{4} & 0.000 & \textbf{4} (5/5) & 0.753 \\
mug100\_1 & 100 & 166 & \textbf{4} & 0.003 & \textbf{4} & 0.000 & \textbf{4} (5/5) & 17.8 \\
mug100\_25 & 100 & 166 & \textbf{4} & 0.003 & \textbf{4} & 0.000 & \textbf{4} (5/5) & 13.0 \\
\midrule
\multicolumn{9}{l}{\textit{mulsol}} \\
mulsol.i.2 & 188 & 3885 & \textbf{31} & 0.028 & \textbf{31} & 0.002 & \textbf{31} (5/5) & 38.8 \\
mulsol.i.3 & 184 & 3916 & \textbf{31} & 0.028 & \textbf{31} & 0.002 & \textbf{31} (5/5) & 46.8 \\
\midrule
\multicolumn{9}{l}{\textit{FullIns}} \\
1-FullIns\_3 & 30 & 100 & \textbf{4} & 0.000 & 5 & 0.000 & \textbf{4} (5/5) & 0.116 \\
1-FullIns\_4 & 93 & 593 & \textbf{5} & 0.004 & 6 & 0.000 & \textbf{5} (5/5) & 19.2 \\
2-FullIns\_3 & 52 & 201 & \textbf{5} & 0.001 & 6 & 0.000 & \textbf{5} (5/5) & 0.567 \\
3-FullIns\_3 & 80 & 346 & \textbf{6} & 0.002 & 7 & 0.000 & \textbf{6} (5/5) & 2.506 \\
\midrule
\multicolumn{9}{l}{\textit{Insertions}} \\
1-Insertions\_4 & 67 & 232 & \textbf{5} & 0.002 & \textbf{5} & 0.000 & \textbf{5} (5/5) & 0.815 \\
2-Insertions\_3 & 37 & 72 & \textbf{4} & 0.001 & \textbf{4} & 0.000 & \textbf{4} (5/5) & 0.164 \\
2-Insertions\_4 & 149 & 541 & \textbf{5} & 0.008 & \textbf{5} & 0.001 & \textbf{5} (5/5) & 5.851 \\
3-Insertions\_3 & 56 & 110 & \textbf{4} & 0.001 & \textbf{4} & 0.000 & \textbf{4} (5/5) & 0.462 \\
3-Insertions\_4 & 281 & 1046 & \textbf{5} & 0.027 & \textbf{5} & 0.002 & \textbf{5} (5/5) & 38.9 \\
\midrule
\multicolumn{9}{l}{\textit{le450}} \\
le450\_5a & 450 & 5714 & 10 & 0.122 & 11 & 0.009 & \textbf{7} (1/5) & 32.9 \\
le450\_5b & 450 & 5734 & 9 & 0.122 & 10 & 0.009 & \textbf{7} (3/5) & 30.7 \\
le450\_5d & 450 & 9757 & 12 & 0.168 & 11 & 0.010 & \textbf{5} (5/5) & 1912.9 \\
le450\_15a & 450 & 8168 & 17 & 0.160 & 17 & 0.010 & \textbf{16} (5/5) & 96.5 \\
le450\_15b & 450 & 8169 & \textbf{16} & 0.161 & \textbf{16} & 0.010 & \textbf{16} (5/5) & 454.9 \\
le450\_15c & 450 & 16680 & \textbf{23} & 0.260 & 24 & 0.012 & \textbf{23} (5/5) & 160.5 \\
le450\_15d & 450 & 16750 & 24 & 0.274 & 26 & 0.013 & \textbf{23} (5/5) & 112.4 \\
le450\_25a & 450 & 8260 & \textbf{25} & 0.173 & 26 & 0.009 & \textbf{25} (5/5) & 1471.9 \\
le450\_25d & 450 & 17425 & 28 & 0.281 & 29 & 0.012 & \textbf{27} (2/5) & 67.0 \\
\midrule
\multicolumn{9}{l}{\textit{school}} \\
school1 & 385 & 19095 & 17 & 0.242 & 15 & 0.010 & \textbf{14} (5/5) & 309.4 \\
school1\_nsh & 352 & 14612 & 27 & 0.166 & 16 & 0.008 & \textbf{15} (5/5) & 164.8 \\
\midrule
\multicolumn{9}{l}{\textit{latin\_square}} \\
latin\_square\_10 & 900 & 307350 & 132 & 7.224 & 134 & 0.085 & \textbf{122} (1/5) & 91.6 \\
\end{longtable}
}

\subsubsection{FAP Instances}

{\footnotesize
\begin{longtable}{lrrrrrrrr}
\caption{FAP instances results.}\label{table:fap-instances} \\
\toprule
& & & \multicolumn{2}{c}{\textbf{DSATUR}} & \multicolumn{2}{c}{\textbf{GISD}} & \multicolumn{2}{c}{\textbf{SSLD}} \\
\cmidrule(lr){4-5} \cmidrule(lr){6-7} \cmidrule(lr){8-9}
\textbf{Instance} & $|V|$ & $|E|$ & $k$ & $t(s)$ & $k$ & $t(s)$ & $k$ (hit) & $t(s)$ \\
\midrule
\endfirsthead
\multicolumn{9}{l}{\small\textit{(continued from previous page)}} \\
\toprule
& & & \multicolumn{2}{c}{\textbf{DSATUR}} & \multicolumn{2}{c}{\textbf{GISD}} & \multicolumn{2}{c}{\textbf{SSLD}} \\
\cmidrule(lr){4-5} \cmidrule(lr){6-7} \cmidrule(lr){8-9}
\textbf{Instance} & $|V|$ & $|E|$ & $k$ & $t(s)$ & $k$ & $t(s)$ & $k$ (hit) & $t(s)$ \\
\midrule
\endhead
\midrule
\multicolumn{9}{r}{\small\textit{continued on next page}} \\
\endfoot
\bottomrule
\endlastfoot
\multicolumn{9}{l}{\textit{CELAR}} \\
CELAR01 & 916 & 5548 & \textbf{12} & 0.338 & 13 & 0.028 & T/O (--) & 3600 \\
CELAR02 & 200 & 1235 & \textbf{13} & 0.017 & \textbf{13} & 0.002 & \textbf{13} (5/5) & 11.1 \\
CELAR03 & 400 & 2760 & 13 & 0.069 & \textbf{12} & 0.006 & \textbf{12} (5/5) & 911.2 \\
CELAR04 & 680 & 3967 & \textbf{13} & 0.190 & \textbf{13} & 0.016 & T/O (--) & 3600 \\
CELAR05 & 400 & 2598 & \textbf{12} & 0.069 & \textbf{12} & 0.006 & T/O (--) & 3600 \\
CELAR06 & 200 & 1322 & \textbf{20} & 0.016 & \textbf{20} & 0.002 & \textbf{20} (5/5) & 14.5 \\
CELAR07 & 400 & 2865 & \textbf{20} & 0.070 & \textbf{20} & 0.006 & \textbf{20} (5/5) & 660.4 \\
CELAR11 & 680 & 4103 & \textbf{20} & 0.197 & \textbf{20} & 0.017 & T/O (--) & 3600 \\
\midrule
\multicolumn{9}{l}{\textit{GRAPH}} \\
GRAPH01 & 200 & 1134 & \textbf{18} & 0.017 & \textbf{18} & 0.002 & \textbf{18} (5/5) & 267.7 \\
GRAPH02 & 400 & 2245 & \textbf{14} & 0.065 & \textbf{14} & 0.006 & T/O (--) & 3600 \\
GRAPH03 & 200 & 1134 & \textbf{12} & 0.015 & \textbf{12} & 0.002 & \textbf{12} (5/5) & 166.7 \\
GRAPH04 & 400 & 2244 & \textbf{14} & 0.064 & \textbf{14} & 0.006 & T/O (--) & 3600 \\
GRAPH05 & 200 & 1134 & \textbf{18} & 0.016 & \textbf{18} & 0.002 & \textbf{18} (5/5) & 262.9 \\
GRAPH08 & 680 & 3757 & \textbf{16} & 0.188 & \textbf{16} & 0.015 & \textbf{16} (5/5) & 1791.0 \\
GRAPH09 & 916 & 5246 & \textbf{18} & 0.358 & \textbf{18} & 0.026 & T/O (--) & 3600 \\
GRAPH14 & 916 & 4638 & \textbf{8} & 0.319 & \textbf{8} & 0.031 & \textbf{8} (5/5) & 116.0 \\
\end{longtable}
}

\subsubsection{Job Shop Scheduling Instances}

{\footnotesize
\begin{longtable}{lrrrrrrrr}
\caption{Job Shop Scheduling instances results.}\label{table:jobshop-instances} \\
\toprule
& & & \multicolumn{2}{c}{\textbf{DSATUR}} & \multicolumn{2}{c}{\textbf{GISD}} & \multicolumn{2}{c}{\textbf{SSLD}} \\
\cmidrule(lr){4-5} \cmidrule(lr){6-7} \cmidrule(lr){8-9}
\textbf{Instance} & $|V|$ & $|E|$ & $k$ & $t(s)$ & $k$ & $t(s)$ & $k$ (hit) & $t(s)$ \\
\midrule
\endfirsthead
\multicolumn{9}{l}{\small\textit{(continued from previous page)}} \\
\toprule
& & & \multicolumn{2}{c}{\textbf{DSATUR}} & \multicolumn{2}{c}{\textbf{GISD}} & \multicolumn{2}{c}{\textbf{SSLD}} \\
\cmidrule(lr){4-5} \cmidrule(lr){6-7} \cmidrule(lr){8-9}
\textbf{Instance} & $|V|$ & $|E|$ & $k$ & $t(s)$ & $k$ & $t(s)$ & $k$ (hit) & $t(s)$ \\
\midrule
\endhead
\midrule
\multicolumn{9}{r}{\small\textit{continued on next page}} \\
\endfoot
\bottomrule
\endlastfoot
\multicolumn{9}{l}{\textit{p-instances}} \\
p06 & 16 & 38 & \textbf{4} & 0.000 & \textbf{4} & 0.000 & \textbf{4} (5/5) & 0.028 \\
p07 & 24 & 92 & \textbf{5} & 0.000 & \textbf{5} & 0.000 & \textbf{5} (5/5) & 0.073 \\
p08 & 24 & 92 & \textbf{5} & 0.000 & \textbf{5} & 0.000 & \textbf{5} (5/5) & 0.073 \\
p09 & 25 & 100 & \textbf{5} & 0.000 & \textbf{5} & 0.000 & \textbf{5} (5/5) & 0.045 \\
p10 & 16 & 32 & \textbf{4} & 0.000 & \textbf{4} & 0.000 & \textbf{4} (5/5) & 0.028 \\
p11 & 18 & 48 & \textbf{5} & 0.000 & \textbf{5} & 0.000 & \textbf{5} (5/5) & 0.035 \\
p12 & 26 & 90 & \textbf{5} & 0.000 & \textbf{5} & 0.000 & \textbf{5} (5/5) & 0.077 \\
p13 & 34 & 160 & \textbf{6} & 0.001 & \textbf{6} & 0.000 & \textbf{6} (5/5) & 0.133 \\
p14 & 31 & 110 & \textbf{6} & 0.001 & \textbf{6} & 0.000 & \textbf{6} (5/5) & 0.070 \\
p15 & 34 & 136 & \textbf{6} & 0.001 & \textbf{6} & 0.000 & \textbf{6} (5/5) & 0.141 \\
p16 & 34 & 134 & \textbf{6} & 0.001 & \textbf{6} & 0.000 & \textbf{6} (5/5) & 0.085 \\
p17 & 37 & 161 & \textbf{7} & 0.001 & \textbf{7} & 0.000 & \textbf{7} (5/5) & 0.100 \\
p18 & 35 & 143 & \textbf{6} & 0.001 & 7 & 0.000 & \textbf{6} (5/5) & 0.087 \\
p19 & 36 & 156 & \textbf{7} & 0.001 & \textbf{7} & 0.000 & \textbf{7} (5/5) & 0.121 \\
p20 & 37 & 142 & \textbf{6} & 0.001 & \textbf{6} & 0.000 & \textbf{6} (5/5) & 0.099 \\
p21 & 38 & 155 & \textbf{7} & 0.001 & \textbf{7} & 0.000 & \textbf{7} (5/5) & 0.103 \\
p22 & 38 & 154 & \textbf{6} & 0.001 & \textbf{6} & 0.000 & \textbf{6} (5/5) & 0.124 \\
p23 & 44 & 204 & \textbf{7} & 0.001 & \textbf{7} & 0.000 & \textbf{7} (5/5) & 0.132 \\
p24 & 34 & 104 & \textbf{6} & 0.001 & \textbf{6} & 0.000 & \textbf{6} (5/5) & 0.092 \\
\midrule
\multicolumn{9}{l}{\textit{r-instances}} \\
r01 & 144 & 1280 & \textbf{13} & 0.010 & \textbf{13} & 0.001 & \textbf{13} (5/5) & 1.533 \\
r05 & 142 & 1266 & \textbf{13} & 0.009 & \textbf{13} & 0.001 & \textbf{13} (5/5) & 1.789 \\
r10 & 150 & 1409 & \textbf{13} & 0.011 & \textbf{13} & 0.001 & \textbf{13} (5/5) & 1.631 \\
r15 & 198 & 2055 & \textbf{16} & 0.019 & \textbf{16} & 0.002 & \textbf{16} (5/5) & 3.711 \\
\midrule
\multicolumn{9}{l}{\textit{GEOM}} \\
GEOM30 & 30 & 50 & \textbf{6} & 0.000 & \textbf{6} & 0.000 & \textbf{6} (5/5) & 0.073 \\
GEOM50 & 50 & 127 & \textbf{6} & 0.001 & \textbf{6} & 0.000 & \textbf{6} (5/5) & 0.616 \\
GEOM100 & 100 & 547 & \textbf{9} & 0.004 & 10 & 0.001 & \textbf{9} (5/5) & 58.5 \\
GEOM110 & 110 & 638 & \textbf{9} & 0.005 & 10 & 0.001 & \textbf{9} (5/5) & 34.0 \\
GEOM120 & 120 & 773 & \textbf{11} & 0.006 & \textbf{11} & 0.001 & \textbf{11} (5/5) & 32.3 \\
\end{longtable}
}

\subsubsection{Random Graphs}

For each random graph family, we generate $n=500$ instances and report how SSLD compares against both DSATUR and GISD. The wins column counts instances where SSLD uses fewer colors than the baseline, ties where it uses the same, and loss where it uses more.
$\Delta_{\text{avg}}$ is the mean color reduction $k_{\text{baseline}} - k_{\text{SSLD}}$ computed over wins only, and $\Delta_{\text{max}}$ is the largest single improvement observed.

\begin{table}[H]
\centering
\footnotesize
\begin{tabular}{lrrrrrrrrrrr}
\toprule
& & \multicolumn{5}{c}{\textbf{vs DSATUR}} & \multicolumn{5}{c}{\textbf{vs GISD}} \\
\cmidrule(lr){3-7} \cmidrule(lr){8-12}
\textbf{Family} & $n$ & \textbf{wins} & $\Delta_{\text{avg}}$ & $\Delta_{\text{max}}$ & \textbf{ties} & \textbf{loss} & \textbf{wins} & $\Delta_{\text{avg}}$ & $\Delta_{\text{max}}$ & \textbf{ties} & \textbf{loss} \\
\midrule
ER & 500 & 416 & 1.84 & 5 & 84 & 0 & 443 & 1.93 & 5 & 57 & 0 \\
BA & 500 & 71 & 1.00 & 1 & 427 & 2 & 98 & 1.00 & 1 & 401 & 1 \\
WS & 500 & 350 & 1.19 & 5 & 150 & 0 & 403 & 1.18 & 5 & 97 & 0 \\
\midrule
\textbf{Total} & \textbf{1500} & \textbf{837} & \textbf{1.50} & \textbf{5} & \textbf{661} & \textbf{2} & \textbf{944} & \textbf{1.51} & \textbf{5} & \textbf{555} & \textbf{1} \\
\bottomrule
\end{tabular}
\caption{}
\end{table}

\subsection{Runtime \& Performance Analysis}\label{sec:runtime}

\begin{figure}[h]
\centering
\includegraphics[width=\textwidth]{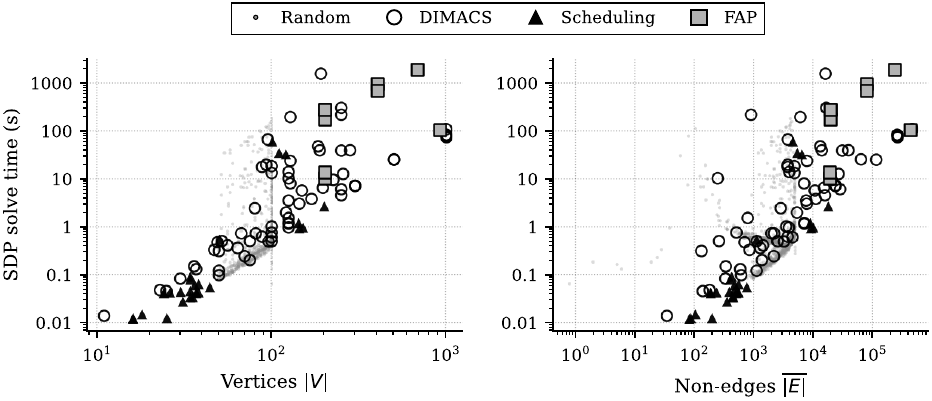}
\caption{SDP solve time as a function of density across benchmark families. One point per instance (1604 in total); axes logarithmic except for density. The SDP solve time is defined per instance.}
\label{sdp_time}
\end{figure}

The SSLD algorithm is decomposed into two phases: one SDP solve and $T$ rounding iterations. Each rounding iteration follows the randomized rounding procedure of Karger, Motwani, and Sudan \cite{karger1998approximate}: a random projection is applied to the SDP embedding, followed by a greedy extraction of an independent set and a DSATUR completion.
We compare their respective runtimes.
Figure~\ref{sdp_time} shows the SDP solve time as a function of the vertex count $|V|$ and the non-edge count across the benchmark families.

The diagonal of the matrix $X$ optimized in the D-SDP is pinned to $1$ and every edge entry to $0$, so the free variables of the program are exactly the non-adjacent pairs.
Solve time grows with both quantities, going from $1.2 \times 10^{-2}$s on the smallest scheduling instances to $1.8 \times 10^3$s on the largest FAP instances.

Order is the strongest structural predictor since the solve time is growing roughly as $|V|^{2.4}$. Because $\overline{|E|}$ grows like $|V|^2$, it tells us nothing beyond $|V|$, so the second panel reports how many free variables the program has rather than an independent cause of its cost.

\begin{figure}[h]
\centering
\includegraphics[width=0.9\textwidth]{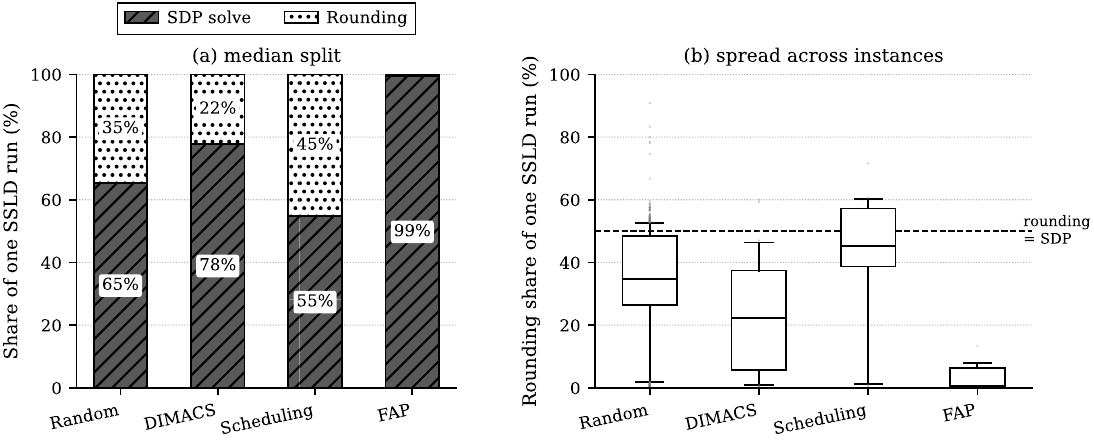}
\caption{Runtime breakdown between the SDP solve and the rounding phase.
(a) median split per family.
(b) distribution of the rounding share across instances, boxes spanning the inter-quartile range with whiskers at the 5th and 95th percentiles. The dashed line marks where the two phases cost the same.}
\label{time_breakdown}
\end{figure}

The rounding phase is not negligible at $T = 500$. Over a single SSLD run, one SDP solve plus one rounding pass, the rounding accounts for a median of $34.7\%$ of total runtime (mean $34.1\%$, 90th percentile $51.2\%$) and it exceeds the SDP on $280$ of $1604$ instances ($17.5\%$). The split is strongly family-dependent, as Figure~\ref{time_breakdown} shows: rounding represents approximately $0.5\%$ of a run on the large FAP instances, $22.3\%$ on DIMACS, $34.8\%$ on the random graphs and $45.3\%$ on the small scheduling instances. The SDP therefore dominates only where it is expensive in absolute terms.

\begin{figure}[h]
\centering
\includegraphics[width=\textwidth]{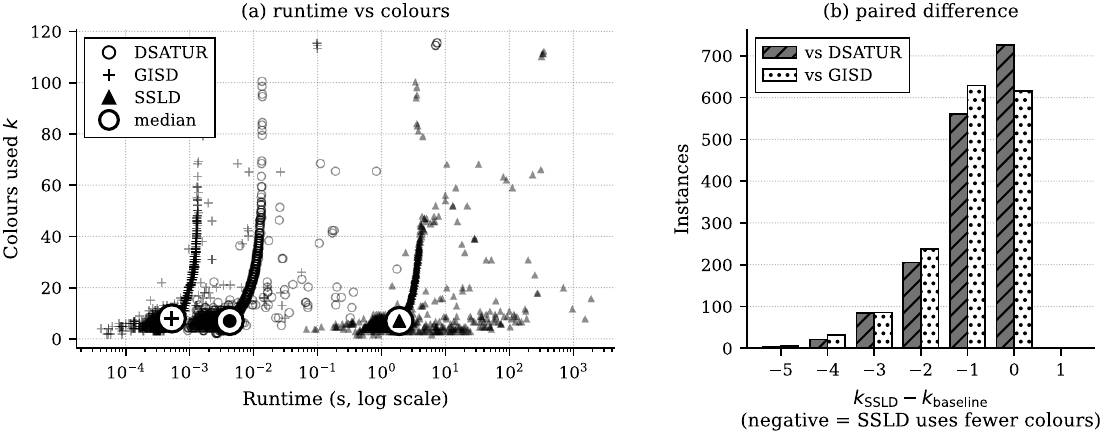}
\caption{Solution quality vs runtime across all benchmark instances.
(a) one point per instance per algorithm; the large outlined marker is the per-algorithm median. Lower and further left is better.
(b) per-instance paired difference in colors between SSLD and each baseline; negative values are instances where SSLD uses fewer colors.}
\label{quality_vs_time}
\end{figure}

Figure~\ref{quality_vs_time} illustrates the trade-off between quality and runtime across all benchmark instances. DSATUR and GISD are faster by two to three orders of magnitude: the median runtime is $5.3 \times 10^{-4}$ s for GISD and $4.2 \times 10^{-3}$ s for DSATUR, against $0.82$ s for SSLD, a factor of $195$ over DSATUR. Panel (a) alone cannot establish the quality claim since $k$ ranges from $2$ to $115$ across instances while the three algorithms differ by well under one color, so their points nearly coincide on the $k$ axis. Panel (b) provides the paired evidence: against DSATUR, SSLD uses fewer colors on $876$ instances, the same number on $726$ instances, and more on $2$ instances, for a mean of $-0.83$ colors; against GISD the record is $988$ / $615$ / $1$, for a mean of $-0.94$ colors. SSLD is thus not intended to replace DSATUR in time-critical settings, but rather to serve as a higher-quality alternative when runtime is not the primary constraint.

\section{Conclusion}

We introduced SSLD, a preprocessing strategy for DSATUR that fixes a single color class chosen via an SDP before completing the coloring with DSATUR.
Across over 1600 benchmark instances considered, through the DIMACS and COLOR challenges, frequency assignment and job shop scheduling instances, the Spinrad--Vijayan adversarial family, and three synthetic random graph models, SSLD matches or improves on plain DSATUR in almost every case.

GISD uses more colors than DSATUR on several instances, confirming that fixing an arbitrary color class first does not consistently help. By contrast, SSLD uses fewer colors than DSATUR on 876 of the 1604 instances tested and fewer than GISD on 988 instances, with only 2 and 1 losses respectively. This confirms that the gain is attributable specifically to the SDP-guided choice of the first color class rather than to the simple act of fixing a color class before using DSATUR.

This improvement in using fewer colors comes at a substantial runtime cost.
SSLD is roughly 195 times slower than DSATUR in median runtime, and this overhead is not confined to the SDP solve itself: the rounding phase that consists of $T = 500$ independent projections and DSATUR completions, accounts for a median of 34.7\% of total runtime and exceeds the cost of the SDP solve on 17.5\% of instances (Section~\ref{sec:runtime}).
Since SSLD's runtime grows with graph size and density, it causes timeouts on several CELAR and GRAPH instances.

The embedding produced by the D-SDP carries no mathematical guarantee that the independent sets extracted from it are optimal or even good as a first color class for DSATUR: while the D-SDP objective encourages alignment of non-adjacent vertices, nothing in the formulation ensures that the resulting random projections will result in an independent set that minimizes the number of colors DSATUR needs to complete the coloring. Nevertheless, there is reason to believe that SDP-based preprocessing can be made theoretically grounded. Karger, Motwani and Sudan \cite{karger1998approximate} did manage to show that an SDP relaxation linked to the Lovász theta function, combined with randomized rounding, results a polynomial-time approximation algorithm for graph coloring with provable guarantees: $O(n^{0.387})$ colors on 3-colorable graphs and $O(n^{1 - 3/(k+1)})$ colors on $k$-colorable graphs. Therefore, it is a future direction to attempt to establish whether the D-SDP embedding can admit approximation guarantees.

Several other directions are worth pursuing to build on this work.
The first one concerns the random projection step, we currently apply a blind search over $T$ attempts with no guidance on what makes a good direction $r$, it could be replaced by a learned or locally-optimized projection, which would help reducing wasted iterations.
More broadly, rather than fixing a single color class before using DSATUR, we could iteratively fix multiple color classes, each guided by the SDP embedding of the residual graph.
The rounding procedure itself could also be made smarter, for instance by combining multiple directions simultaneously or by biasing the projection using the previously obtained coloring.

Finally, the SDP solve remains the dominant runtime cost on large instances; warm starting, exploiting sparsity, or replacing the SDP with a cheaper relaxation are potential directions to make SSLD practical on denser and larger graphs.

\bibliographystyle{ieeetr}
\bibliography{ref}
\end{document}